%% file: acl_latex_v2.tex
\pdfoutput=1

\documentclass[11pt]{article}

\usepackage{acl}

\usepackage{times}
\usepackage{latexsym}
\usepackage{booktabs}
\usepackage{multirow}

\usepackage[T1]{fontenc}

\usepackage[utf8]{inputenc}

\usepackage{microtype}
\usepackage{geometry}
\usepackage{array}
\usepackage{gnuplottex}
\usepackage{lipsum}
\usepackage{graphicx}
\usepackage{tikz}
\usepackage{pgfplots}
\usepackage{xspace}
\usepackage{enumitem}
\usepackage{hyperref}
\usepackage{xurl}

\newcommand{\iclTargetAgnostic}{Target-agnostic\xspace}
\newcommand{\iclTargetSpecific}{Target-centric\xspace}
\newcommand{\iclTargetSpecificPlusWindow}{Target-centric + Context Window\xspace}

\newcommand{\iclTargetAgnosticShort}{Target-agnostic\xspace}
\newcommand{\iclTargetSpecificShort}{Target-centric\xspace}
\newcommand{\iclTargetSpecificPlusWindowShort}{Target-centric + CW\xspace}

\title{Discourse-Aware In-Context Learning for Temporal Expression Normalization}

\author{
Akash Kumar Gautam$^{1,3\thanks{* Work done while the author was an intern at Bosch Center for Artificial Intelligence, Renningen, Germany}}$ \
Lukas Lange$^{1}$ \
Jannik Str\"{o}tgen$^{2}$ \\
  $^1$Bosch Center for Artificial Intelligence, Renningen, Germany \\
  $^2$Karlsruhe University of Applied Sciences, Germany \\
  $^3$ Saarland University, Saarland Informatics Campus, Germany \\
  \texttt{akga00001@stud.uni-saarland.de} \\
  \texttt{Lukas.Lange@de.bosch.com} 
  }

\begin{document}
\maketitle
\begin{abstract}
Temporal expression (TE) normalization is a well-studied problem. However, the predominately used rule-based systems are highly restricted to specific settings, and upcoming machine learning approaches suffer from a lack of labeled data. 
In this work, we explore the feasibility of proprietary and open-source large language models (LLMs) for TE normalization using in-context learning to inject task, document, and example information into the model. 
We explore various sample selection strategies to retrieve the most relevant set of examples. 
By using a window-based prompt design approach, we can perform TE normalization across sentences, while leveraging the LLM knowledge without training the model.
Our experiments show competitive results to models designed for this task. In particular, our method achieves large performance improvements for non-standard settings by dynamically including relevant examples during inference.
\end{abstract}

\section{Introduction}
Temporal tagging is a challenging problem for building information extraction pipelines. Traditionally, it involves first the identification of temporal expressions (TEs) from text (\textbf{extraction}), followed by a mapping to a well-defined format such as TimeML (\textbf{normalization}). Previously, approaches to dealing with this problem involved curating handwritten rules \cite{chang2012sutime, strotgen2013multilingual} often limiting their applicability to new domains and new languages.

\begin{figure}
    \centering
    \includegraphics[width=0.95\linewidth]{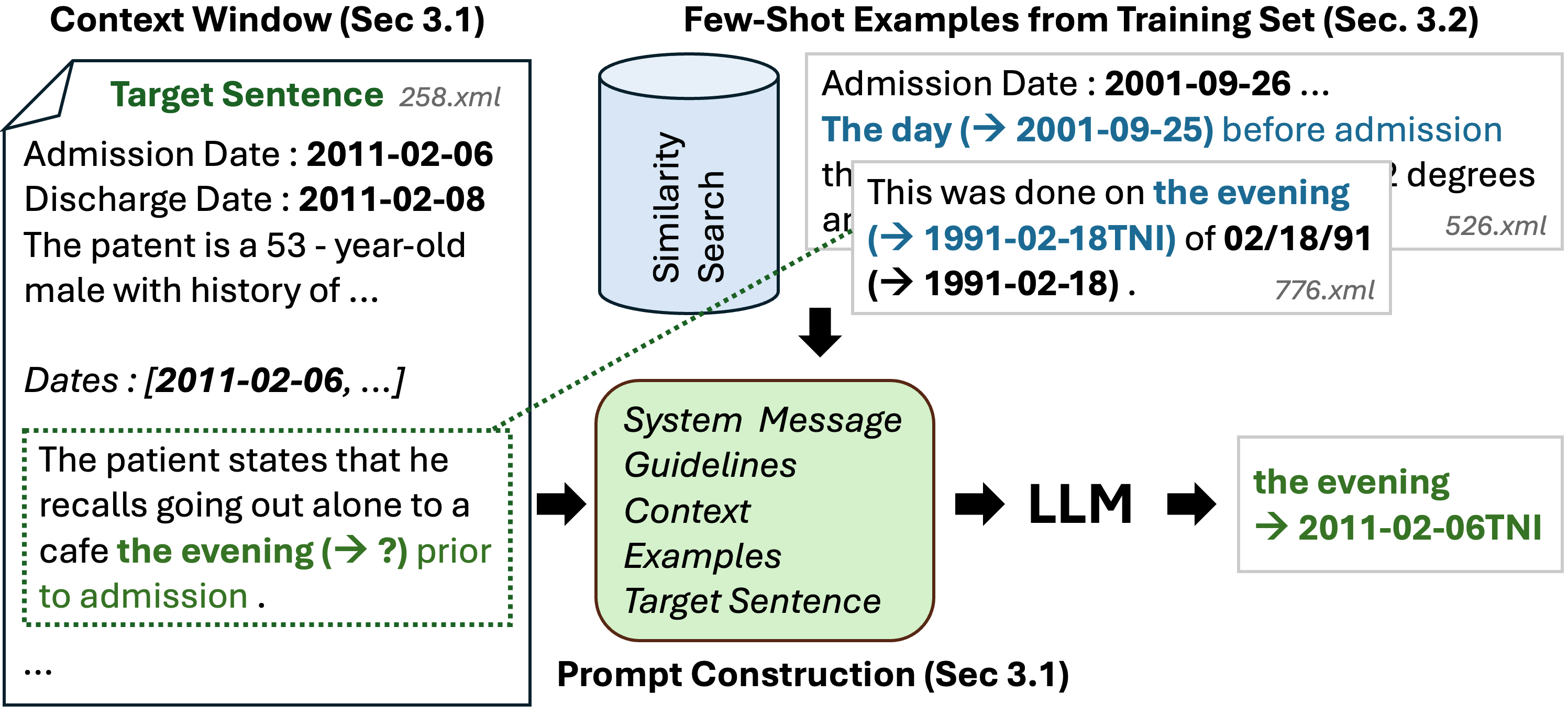}
    \caption{Overview of our proposed in-context learning approach for temporal expression normalization. Given a test input, we retrieve similar text representations from the train set. We combine both of them along with a running context window of previous predictions and feed it to a language model along with instructions.}
    \label{fig:main_fig}
\end{figure}

More recently, deep neural networks were trained for many tasks across domains and languages \cite{rahimi2019massively, artetxe2019massively}. However, they require an increasing amount of training data. In contrast, the advent of large language models (LLMs) \cite{brown2020language,rae2022scaling} led to strong zero- and few-shot capabilities by transferring knowledge for specific downstream NLP tasks like named entity recognition, question-answering, or sequence classification. 
Therefore, making use of a recent LLM without training is a compelling strategy to deal with data scarcity in multilingual setups and also to diversify utility across multiple domains.

In this work, we explore the proprietary GPT-3.5-turbo model as well as the open-source Zephyr model~\citep{tunstall2023zephyr} for TE normalization. For both models, our discourse-aware approach (see Figure~\ref{fig:main_fig}) leverages in-context learning using few-shot examples and a document-level temporal context window. 
We explore various sample selection strategies for prompting tailored toward the TE normalization task and show that standard sentence-level processing might not be suitable to capture all the necessary long-range context dependencies and discourse information.
Our broad evaluation across six domains and seven languages demonstrates the competitiveness of our method to dedicated normalization models. In particular, our analysis reveals the benefits of our method in settings when the target document is more distant from their training data. 

\section{Background and Related Work}

\input{result_table}

\textbf{Temporal Tagging} is a two-step process consisting of TE extraction from textual documents, followed by the normalization into a standard format. We follow the TimeML annotation guidelines \cite{pustejovsky2005specification}, which define four temporal types, namely \textsc{Date}, \textsc{Time}, \textsc{Duration}, and \textsc{Set}. For the normalization, we focus on the \textsc{Value} attribute that captures the most important temporal semantics of a TE.
While explicit TEs include all necessary information for the normalization, e.g., ``May 24, 2024'', 
others require further knowledge or temporal discourse information.
For example, implicit expressions like ``Easter 2024'' require semantic knowledge, and relative expressions like ``tomorrow'' rely on an anchoring date, e.g., the Document Creation Time (DCT). Under-specified expressions are missing the relation to the anchor and cannot be fully normalized by the given context.\footnote{For further details, we refer to \cite{strotgen2016domain}.}

\textbf{TE Extraction and Reasoning.}
TE extraction has been handled as a sequence labeling problem through trained language model sequence taggers \cite{laparra2018characters, lange2020adversarial}. 
\citet{lin2019bert} utilize BERT to identify temporal relations in text.
\citet{chu2023timebench} investigate temporal reasoning capabilities of recent LLMs. In extraction-related tasks, prior works explore GPT's abilities for event extraction, specifically relation extraction \cite{tang2023does, gao2023exploring, wei2023zeroshot}. However, no prior work studies the feasibility of TE normalization using LLMs. 

For solving TE normalization, several rule-based systems have been proposed such as HeidelTime~\citep{strotgen2013multilingual} and SUTime~\citep{chang2012sutime}, while other systems rely on context-free grammars \cite{bethard2013synchronous, lee2014context}. However, both approaches rely on highly language-specific resources. In contrast, deep-learning-based models have demonstrated robust and generalizable performance across languages for the normalization \citep{lange2023multilingual}. However, this system required a careful design of the neural network and large-scale training. Instead, we rely on the transfer learning abilities of LLMs by providing selected examples to learn from. 

\textbf{Few-Shot Learning.}
With the advent of powerful pre-trained language models, \citet{brown2020language} discovered that these models can be utilized for solving tasks without task-specific training. By providing examples and task descriptions, these models can generalize their existing knowledge and transfer this to follow the given instructions. 
Common approaches involve passing representative examples from the training set, through manual or automatic selection strategies, in task-specific formats to the LLM \cite{min2022metaicl,rubin2022learning}. Successful approaches are based on paraphrasing methods where initial text seed prompts are paraphrased into semantically similar expressions, with a further combination involving criteria like Maximal Marginal Relevance \cite{mao2020multi}.
As the context length of LLMs is limited and commercial APIs charge per input token, the selection criteria for sample selection becomes a crucial factor for the performance and applicability.

\section{Approach}
In-context learning (ICL) utilizes few-shot examples to learn the downstream task \cite{min2022metaicl}. 
We follow this approach and describe our selection strategies along with prompting formats relevant to TE normalization.

\subsection{Prompt Format}\label{sec:prompt}
We follow the best practices for LLM decoding and regulating the output behavior by defining various prompt inputs. Sample prompts are given in Appendix \ref{app:prompt} that showcase our prompt structure. In general, we provide information on the task, the document context, a selected set of samples, and the expected JSON output format.

\label{sec:window}
For the context, we process all sentences from a document $d$ containing temporal expressions (target sentences) sequentially, as later temporal expressions might need earlier seen temporal expressions as reference times. 
For this, we maintain a running record of previously seen TEs from $d$ to support the anchoring of non-explicit expressions to the correct temporal scope, including the DCT.
These running records of previously seen TEs serve as temporal containers that will allow the model to have enough semantic information to normalize relative or under-specific expressions correctly\cite{strotgen2016domain}. 

Given a target sentence $t$ containing one or more TEs, 
we aim to provide similar sentences
with TEs from the candidate pool as few-shot examples. These examples are retrieved from the respective training corpora and should enable the LLM in-context learning to normalize the TEs in $t$.

\subsection{Few-Shot Example Selection}
We now describe our selection strategies for the few-shot examples. For this, we use semantic search to select samples from the training sets given a target sentence $t$. In all setups, we use the 
embedding model\footnote{\url{https://huggingface.co/intfloat/multilingual-e5-base}} to create vector representations of text sequences and select examples based on the embedding similarity between candidate sentences and $t$.

\textbf{\iclTargetAgnostic.} The \textit{k} most dissimilar examples from the training set are selected, as random sampling can lead to clusters of similar sentences. With this, we want to create a diverse and representative set 
that is useful for all target sentences.

\textbf{\iclTargetSpecific}. The \textit{k} most similar sentences or documents are selected given the target sentence or document. Selecting entire documents might allow the model to better learn long-term dependencies.

\textbf{\iclTargetSpecificPlusWindow}. As the LLM input length is limited, we restrict the normalization to a single sentence at a time.
This allows to increase the number of selected samples without compromising the performance. To capture long-term temporal dependencies for TEs, 
we record previously processed sentences of the same document as a fixed-length context window 
(see Section~\ref{sec:window}).\footnote{We use the predicted \textsc{Value} attributes as context.}

\textbf{Expert Prompt.} We experiment with examples derived from the TimeML guidelines. We assume that these are representative enough for the model to understand the task and the normalization format. The full prompt is given in Appendix \ref{app:prompt}.

\section{Experiments}
This section describes our experimental setup, the results, and broad analyses of various settings.

\textbf{Data.} For our experiments, we use 4 English datasets from various domains to evaluate the generalizability of our approach. 
This includes the popular TempEval3 \citep{uzzaman-etal-2013-semeval} (news style), i2b2 \citep{sun2012-i2b2} (clinical), AncientTimes \citep{strotgen-etal-2014-extending} (historical text) and TempCourt \citep{navas-2019-tempcourt} (court decisions) datasets. The latter can be split into three subdomains, depending on the document's origin.\footnote{European Court of Justice (ECJ), United States Supreme Court (USC), European Court of Human Rights (ECHR).} 
We study multilingual in-context learning with AncientTimes resources from six languages: Arabic, Dutch, French, German, Spanish, and Vietnamese. We report average accuracy across 3 different runs as the evaluation metric for all normalization experiments and use the Temp\-Eval3 
evaluation script for temporal tagging setups.

\textbf{Models.}
We experiment with the proprietary GPT-3.5-turbo\footnote{\url{https://platform.openai.com/docs/models/gpt-3-5}} 
and the open-source Zephyr model,\footnote{\url{https://huggingface.co/HuggingFaceH4/zephyr-7b-beta}} which is considered the best-performing open-source 
7B-parameter 
model at the time of writing \footnote{We provide results with other language models in Appendix \ref{app:analyzes}}. 
The maximum input lengths
are 16K and 4K tokens, respectively. 

Since we model TE normalization as a text completion task, we set the temperature parameter to 0 to reduce randomness in the results. All other parameters are kept at their default values. The final input consists of four distinct types of prompts as described in Section~\ref{sec:prompt}. The context window length 
is set to 3, as it showed the highest performance in our initial experiments.

\begin{figure}[t]
    \centering
    \includegraphics[width=1\linewidth]{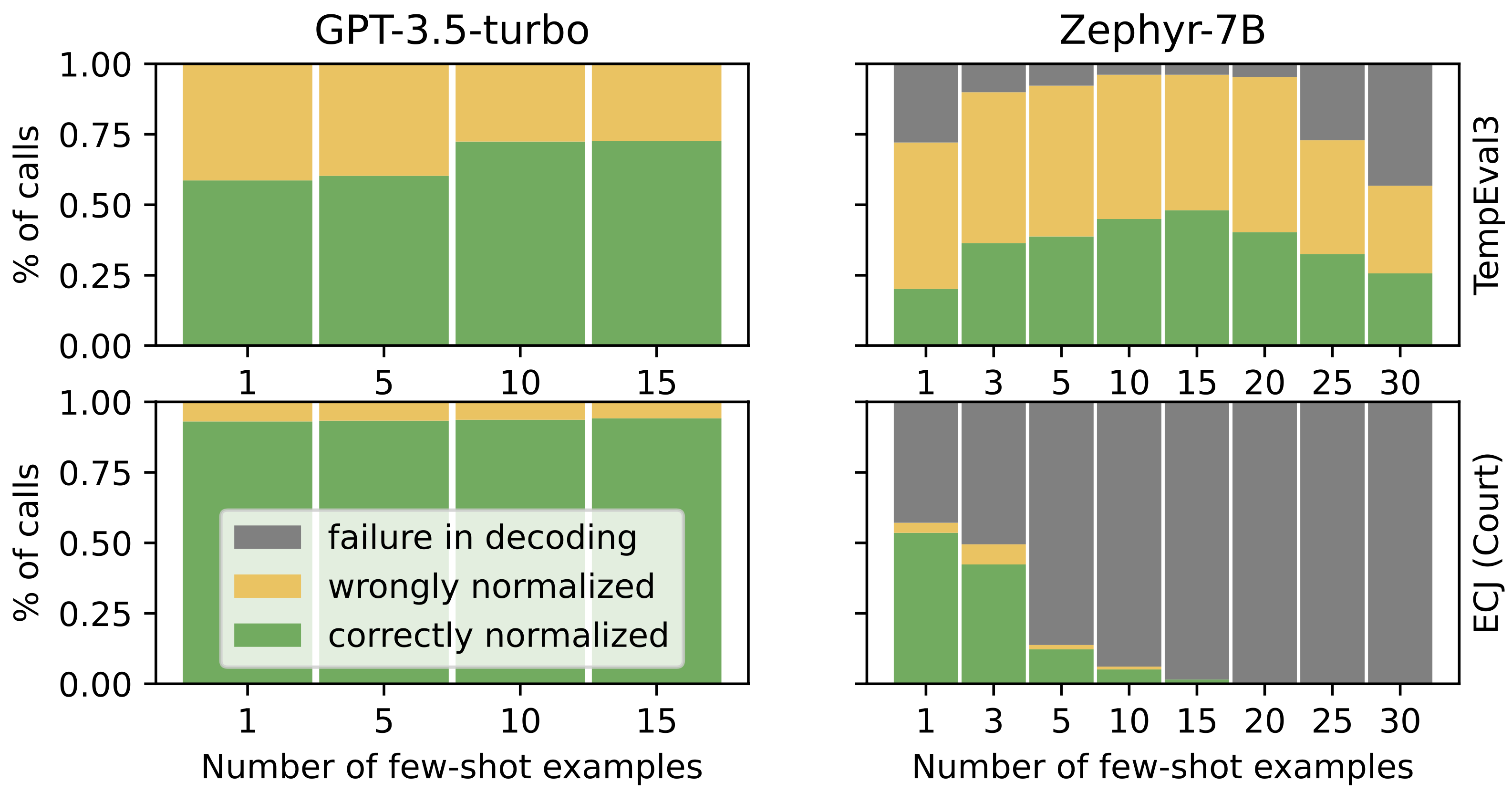}
    \vspace{-0.60cm}
    \caption{Analysis of how the number of examples influences the correctness and failures, e.g., for when the examples exceed the limited context length.
    }
    \label{fig:study_examples}
\end{figure}

\subsection{Results}
\label{sec:results}
The results of our different sample selection strategies are provided in Table~\ref{tab:main_results}. 
The necessity of thoroughly selected few-shot examples 
is emphasized across all datasets, as these methods outperform the expert prompt by a large margin. In particular, the \iclTargetSpecificPlusWindow approach delivers the best ICL performance for five out of six datasets. All of these datasets have a large share of explicit expressions that benefit more from additional examples than document-length context. In contrast, the narrative AncientTimes has dependencies between TEs that can be effectively dealt with only when entire documents are used. This emphasizes that the ICL method should be chosen according to the documents' characteristics. 

The GPT model achieves comparable results to the MLM baseline \citep{lange2023multilingual}, 
except for the clinical i2b2 corpus. In this setting, the target dataset is most distant from the training data of the MLM model, whereas our ICL methods benefit from domain-specific examples. 
The GPT model also considerably outperforms the smaller open-source Zephyr model, which only achieves good performance on the simplest ECHR dataset. Nonetheless, 
this shows the prospects of ICL for complex tasks and also for smaller models.

\subsection{Analysis}
\label{sec:analysis}
We now study different aspects of our method in more detail, i.e., the effects of varying number of few-show examples and different context window lengths. 
We further investigate the application of our method in multilingual setups and in temporal tagging pipelines.

\textbf{Number of Few-Shot Examples.} 
As shown in Figure~\ref{fig:study_examples}, both LLMs reach their performance peak with 10 or 15 examples for the TempEval3 corpus. 
However, the Zephyr model cannot benefit from more examples for the longer ECHR documents. Here, we noticed two failure types for the Zephyr model: (1) The LLM does not output machine-readable JSON, when there are not enough examples to learn the output format. (2) The model exceeds the context lengths with an increasing number of examples. This is partly due to the model's inability to follow the instructions and learn due to limited input context length.

\begin{figure}[t]
    \centering
    \includegraphics[width=0.95\linewidth]{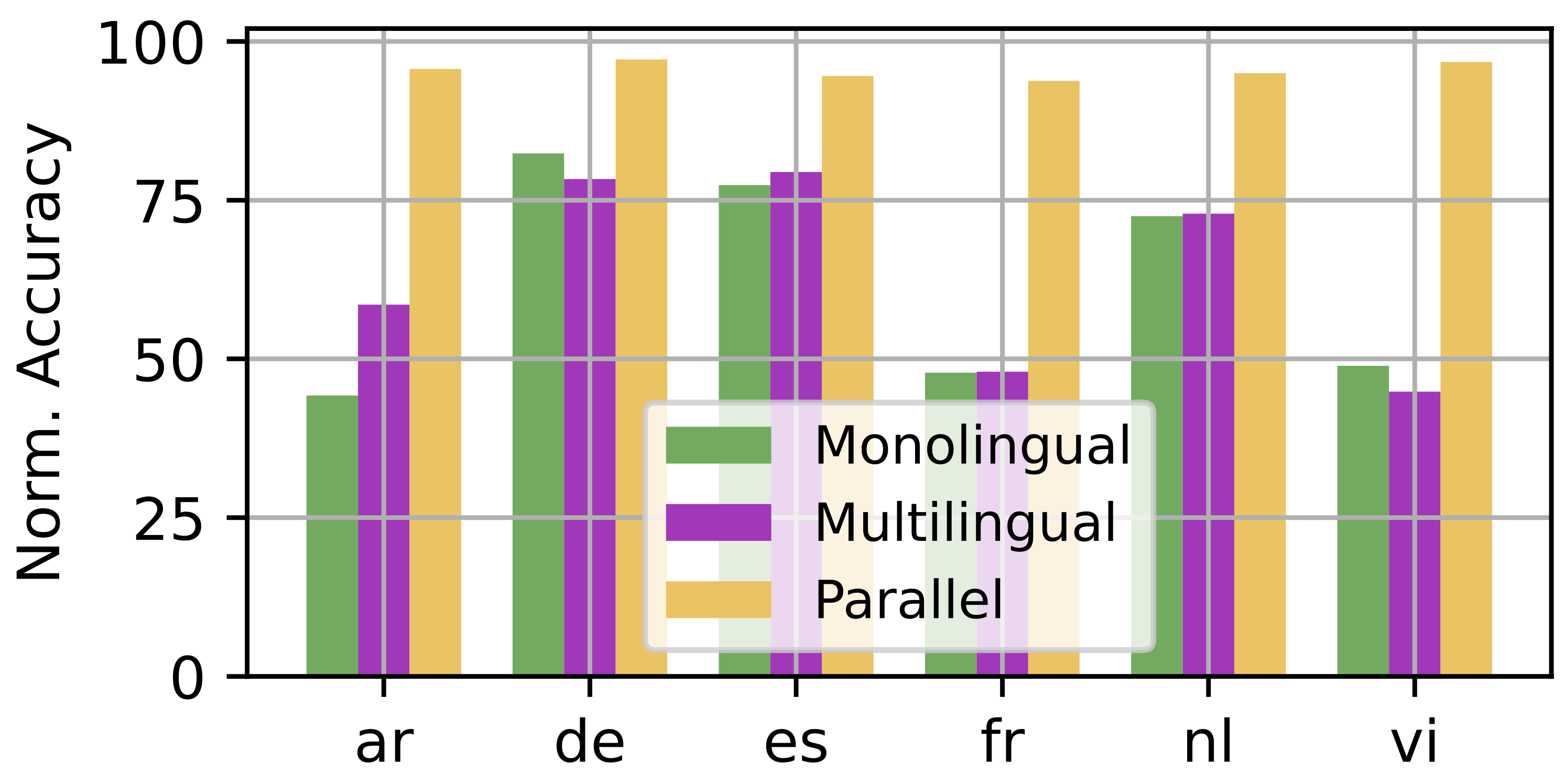}
    \vspace{-0.35cm}
    \caption{Performance on multilingual Ancient-Times corpora with three different sample selection pools.}
    \label{fig:study_multilingual}
\end{figure}

\textbf{Multilingual In-Context Learning.} 
To evaluate if the GPT model can generalize from multilingual examples, we study the effect of our method 
in 3 different settings on the multilingual AncientTimes corpus. \textit{Monolingual:} For each language, we pick same-language samples from the training set. \textit{Multilingual:} We choose samples across languages from the combined training sets of all languages. \textit{Parallel:}
Examples were taken from the train and the test split of all languages, except the target language. The results are given in Figure \ref{fig:study_multilingual}. The general trend suggests that multilingual samples can improve performance, while the highest gain is observed with parallel data. This emphasizes that LLMs can be used for multiple languages without creating language-specific resources, e.g., by translating existing resources.

\begin{figure}[!htbp]
    \centering
    \includegraphics[width=0.7\linewidth]{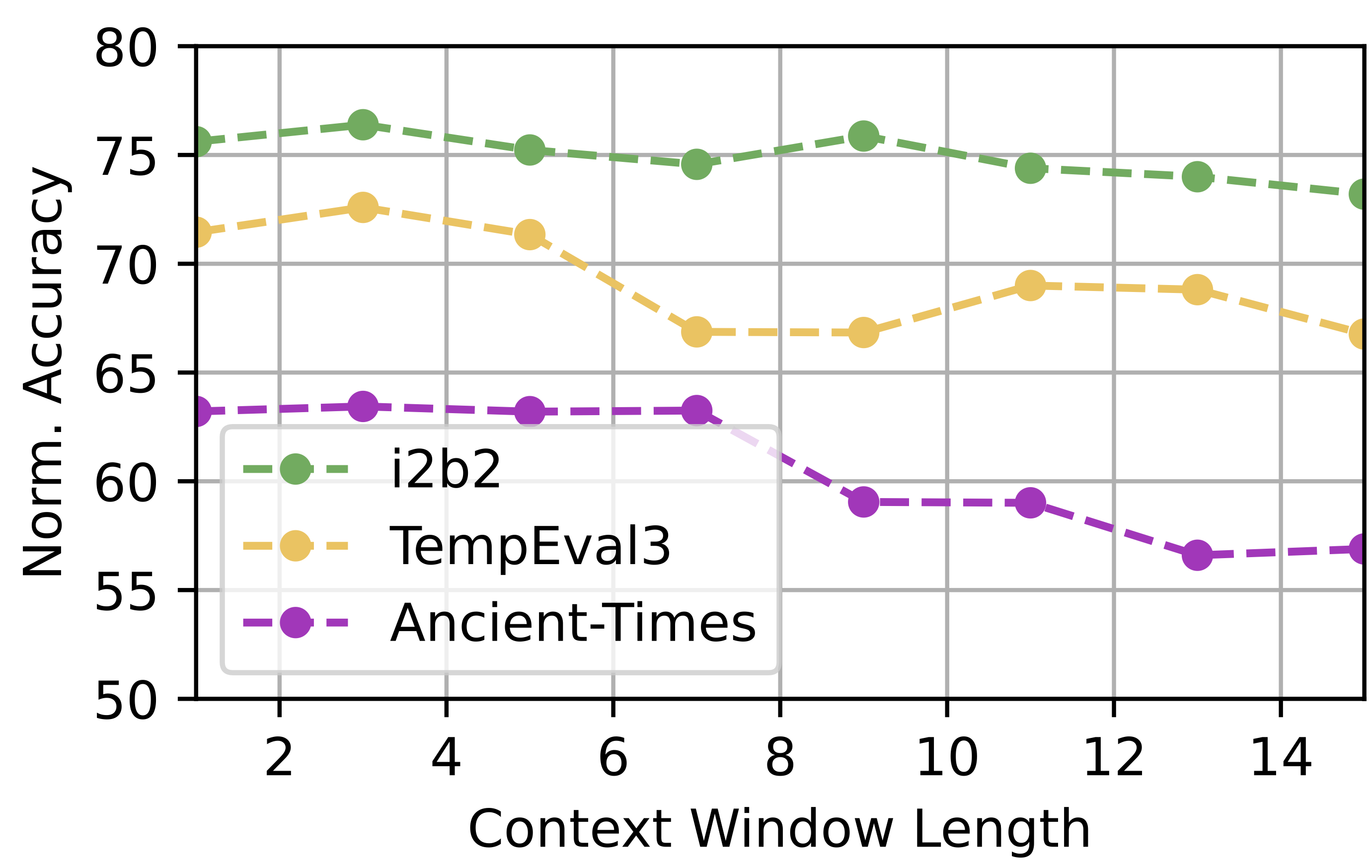}
    \caption{Effect of different context window lengths for our \iclTargetSpecificPlusWindow approach on 3 different corpora.}
    \label{fig:study_context}
\end{figure}

\textbf{Application in a Temporal Tagging System.}
We couple our method with an extraction model, i.e., the NER extraction model from \citep{lange2023multilingual}, to perform full temporal tagging. For this, we train a domain-adapted version of the sequence tagging model to match our data sources. The results are given in Table~\ref{tab:extract} and demonstrate the applicability of our method on extractions from real systems.
As a baseline, we tried to use our ICL method for the extraction such that we could utilize the GPT model as an end-to-end system. However, the poor recall of temporal expressions massively limits the performance of this approach, and ICL for TE extraction would have to be tackled as a separate research question.

\begin{table}[!htpb]
    \centering
    \setlength\tabcolsep{3.5pt}
    \small
    \scalebox{0.9}{
    \begin{tabular}{lcc|cc}
    \toprule
                         & \multicolumn{2}{c|}{\bf TempEval3}     & \multicolumn{2}{c}{\bf  ECJ (Court)}       \\
                         & Extract. & Norm. & Extract. & Norm.  \\
    \midrule
    HeidelTime {\tiny (\citeauthor{strotgen2013multilingual})}
    & 84.1 & 80.0 & 43.3 & 43.0 \\
    NER+MLM {\tiny (\citeauthor{lange2023multilingual})}   
    & 82.8 & 70.5 & 69.5 & 66.0 \\
    \textbf{GPT-ICL} (end-to-end)               
    & 52.1 & 40.5 & 24.6 & 18.2 \\
    \midrule
    Domain-NER+MLM                              
    & \multirow{2}{*}{91.5} & 74.5 & \multirow{2}{*}{94.5} & 90.8 \\
    Domain-NER+\textbf{GPT-ICL}                 
    & & 81.2 & & 91.1 \\
    \bottomrule
    \end{tabular}
    }
    \caption{Application of different normalization methods in real-world extraction+normalization settings.}
    \label{tab:extract}
\end{table}

\textbf{Effect of Context Window Length.} 
Figure \ref{fig:study_context} studies context lengths for three datasets, where the best results are obtained with context lengths between one and five sentences. For longer context sizes, we observe a decrease in performance. This suggests that shorter contexts are often sufficient for LLMs to resolve temporal dependencies. 
Note that the AncientTimes corpus, which benefits from document-level context, does not benefit from an increased context window. We assume that the studied window size may still be too limited for the long-distance dependencies in this setting.

\begin{table}
\centering
\scalebox{0.80}{
\begin{tabular}{lccccc}
\toprule
& {\small \textbf{Explicit}} & {\small \textbf{Implicit}} & {\small \textbf{Relative}} & {\small \textbf{Under-specified}} & {\small \textbf{Vague}} \\
\midrule
{\small Correct} & 38 & 17 & 23 & 05 & 13 \\
{\small False} & 04 & 03 & 08 & 04 & 00 \\
\bottomrule
\end{tabular}
}
\caption{Error analysis w.r.t. different realizations of TEs on examples from the TempEval3 corpus on our approach.}
\label{error_analysis}
\end{table}

\textbf{Error Analysis.} We conduct a manual error inspection of 115 TEs from the TempEval3 corpus regarding their realizations as defined by \citep{strotgen2016domain} plus vague references like ``now'' which is normalized to PRESENT\_REF. The results are provided in Table \ref{error_analysis} and show that our method is able to correctly normalize most explicit, vague, and implicit expressions. The latter benefit from the world knowledge in the LLM. Most challenging are relative and under-specified expressions, where the model lacks enough context or fails to incorporate context information.

\section{Conclusions}
In this paper, we demonstrated that recent LLMs are capable of temporal expression normalization when being prompted with an appropriate in-context learning method. Our discourse-aware prompt allows the LLM to capture important context information while still being generic enough to provide general task descriptions. 
Our experiments across domains and languages showcase the competitive performance of our method compared to specifically designed normalization models and outperforms them when the target document is more distant from their underlying training data.

\section*{Limitations}

Our experiments were limited to seven languages and our insights may not hold for untested languages. Recent advanced literature on example selection strategies presents promising avenues to impart temporal reasoning abilities for improved TE normalization in the ever-growing zoo of LLMs.

\bibliography{anthology,custom}
\bibliographystyle{acl_natbib}

\appendix

\clearpage

\section{Further Analysis}\label{app:analyzes}

\input{results_appendix}

In this section, we share the results of other 7B-parameter language models that we explored for the task of TE normalization. We present the results with Llama2 \citep{touvron2023llama}, Mistral \citep{jiang2024mixtral}, NeuralTrix \footnote{\url{https://huggingface.co/CultriX/NeuralTrix-7B-dpo}}, and Westlake \footnote{\url{https://huggingface.co/senseable/Westlake-7B}}. We observe a large performance gap for all of these models in comparison to the Zephyr model. Upon manual error inspection, we found severe problems regarding their ability to follow instructions, and therefore, to produce valid json outputs.

\section{Implementation Details}
We used Faiss\footnote{\url{https://github.com/facebookresearch/faiss}} to index and cluster vector representations of text sequences (sentences or documents) throughout this work. Spacy\footnote{\url{https://github.com/explosion/spaCy}} was used to split into sentences. The dissimilarity threshold value for the Target-agnostic approach was set to $0.7$. For GPT-3.5-turbo, we also ensure that \texttt{system\_fingerprint} field was consistent across all experiments for the online API calls.\footnote{\url{https://platform.openai.com/docs/guides/text-generation/reproducible-outputs}}

\section{Detailed Prompt Information}\label{app:prompt}
To enable the dynamic and conversational abilities of GPT-3.5-turbo, we make use of \texttt{messages}\footnote{\url{https://community.openai.com/t/on-using-the-messages-array-with-gpt-3-5-turbo-and-gpt-4/367376}} that include further information on how the output and response should be produced. These are intended to 
pass enough context for the conversation model to understand the nuances of the task.

Figure \ref{fig:prompt example} includes a prompt example passed for Target-centric + CW approach for a sample document from the test set. 

We now describe the different types of prompt components, that make up the final API call.

\texttt{SYSTEM PROMPT:} Used to provide system-level instructions to guide the model's behavior throughout the conversation. \\

\texttt{USER PROMPT:} Used to specify the user role for the text input. \\

\texttt{ASSISTANT PROMPT:} Instructions on how the model should respond to the user-level instructions. \\

\texttt{GUIDELINES PROMPT:} Consists of actual text sequences in TimeML annotation format from the train (few-shot examples) and test (target sentence) splits.

\subsection{Expert Prompt}
Figure \ref{fig:expert_prompt} includes the text sequences that were passed as guidelines prompt for the expert prompt example selection strategy mentioned in Section \ref{sec:prompt}.

\begin{figure*}[!htpb]
\centering
\scalebox{0.9}{
    \parbox{\dimexpr\linewidth-2\fboxsep-2\fboxrule}{
        \fbox{
            \begin{minipage}{\dimexpr\linewidth-2\fboxsep-2\fboxrule}
                \textbf{SYSTEM PROMPT:} Function as a system that gives the normalized time expressions for all TIMEX3 tags of type DATE, TIME, DURATION, and SET. 
The identified normalized time expression should be according to TIMEML annotation standards.
The output shows the normalized values for the time expressions. All time expressions that are required to be normalized 
is passed as a list. \\
            
            \textbf{USER PROMPT:} Are you clear about your role? \\

            \textbf{ASSISTANT PROMPT:} Sure, I'm ready to help you with your task. Please provide me with the necessary information to get started. \\
            
                \textbf{GUIDELINES PROMPT:} Here are some examples and the expected output format with normalized expressions
                \begin{enumerate}
                    \item She will need to continue for at least \texttt{<TIMEX3 tid="t17" type="DURATION" previous\_timex="2002-02-01 2002-02-08" dct="2002-02-01">10 more days</TIMEX3>} or as clinically indicated by the course of her cellulitis. \\
                          List of time expressions to normalize: \texttt{['10 more days']} \\
                          Output: \texttt{\{'10 more days': 'P10D'\}}
                    \item Sentence: \textcolor{red}{The patient did well and her suprapubic tube was clamped starting on \texttt{<TIMEX3 tid="t10" type="DATE" value="1993-07-13">postoperative day four</TIMEX3>}. Clamping continued until \texttt{<TIMEX3 tid="t11" type="DATE" value="1993-07-15">postoperative day six</TIMEX3>}. By \texttt{<TIMEX3 tid="t12" type="DATE" value="1993-07-15">postoperative day six</TIMEX3>}, the patient was also tolerating a regular diet and passing flatus.} \textcolor{blue}{In addition, she will take Ciprofloxacin for \texttt{<TIMEX3 tid="t13" type="DURATION" previous\_timex="1993-07-13 1993-07-15" dct="1993-07-09">nine more days</TIMEX3>}}. \\
                          List of time expressions to normalize: \texttt{['nine more days']} 
                          \\
                    Output: {} \\
                \end{enumerate}
            \end{minipage}
        }
        \caption{Prompt Example passed to GPT-3.5 for Target-centric + CW (context window) approach. In the guidelines prompt, sentence \#1 is the text sequence picked from the train set. Sentence \#2 includes text sequences from the test set. Text highlighted in blue is the target sentence passed to the LLM model for normalization. Ones marked in red, are part of the running context window (previous sentences in the same document from the test set, where the VALUE attribute is replaced by predictions from the model.)}
        \label{fig:prompt example}
    }
    }
\end{figure*}

\begin{figure*}[!htpb]
\centering
    \scalebox{0.90}{ 
        \parbox{\dimexpr\linewidth-2\fboxsep-2\fboxrule}{
            \fbox{
                \begin{minipage}{\dimexpr\linewidth-2\fboxsep-2\fboxrule}
                    \textbf{GUIDELINES PROMPT:}
                    \begin{enumerate}
                        \item Reference for ruling visas would be given on \texttt{<TIMEX3 type="DATE" tid="t2">30 April 2013</TIMEX3>}. Written regards to further procedures would be made public on \texttt{<TIMEX3 type="DATE" tid="t3">8 May 2014</TIMEX3>}. \texttt{<TIMEX3 type="DATE" tid="t4">The following day</TIMEX3>} the house will open for discussion. The ceremony for delegates on current immigration laws are held \texttt{<TIMEX3 type="SET" tid="t5">annually</TIMEX3>}. Such kinds of meetings usually lasts only \texttt{<TIMEX3 type="TIME" tid="t6">30 minutes</TIMEX3>}. Such meetings have been going on now for \texttt{<TIMEX3 type="DURATION" tid="t7">more than five years</TIMEX3>} now. Mr. Mark filed for an extension just \texttt{<TIMEX3 type="DURATION" tid="t7">30 days</TIMEX3>} before the expiry of his credentials. \texttt{<TIMEX3 tid="t8" type="DATE">previously</TIMEX3>} he did not do it. In \texttt{<TIMEX3 type="DATE" tid="t9">2016</TIMEX3>} last such case occurred. \\
                        \\
                        List of time expressions to normalize: \texttt{['30 April 2013', '8 May 2014', 'the following day', 'annually', '30 minutes', 'more than five years', '30 days', '2016', 'previously']}
                        \\
                        Output: \texttt{\{'30 April 2013': '2013-04-30', '8 May 2014': '2014-05-08', 'the following day': '2014-05-09', 'annually': 'P1Y', '30 minutes': 'PT30M', 'more than five years': 'P5Y', '30 days': 'P30D', '2016': '2016', 'previously': 'PAST\_REF'\}}
                        
                        \item Sentence: {}
                        
                        Output: 
                    \end{enumerate}
                \end{minipage}
            }
            \caption{Text sequences that were passed as expert prompt example selection strategy. Sequence in Sentence \#2 would be the one from the test set.}
            \label{fig:expert_prompt}
        }
    }
\end{figure*}

\end{document}

%% file: result_table.tex
\begin{table*}[htbp]
    \centering
    \setlength\tabcolsep{6pt}
    \small
    \scalebox{1.0}{
    \begin{tabular}{lllllll}
        \toprule
        & \textbf{Ancient-Times} & \textbf{ECHR} & \textbf{ECJ} & \textbf{USC} & \textbf{i2b2} & \textbf{TempEval3} \\
        Domain & Wikipedia & Court & Court & Court & Clinical & News \\
        \midrule
        MLM \citep{lange2023multilingual}  & \underline{77.0}      & \underline{98.2}      & 93.5      & 86.8      & 48.1      & \underline{79.0} \\
        \midrule
        GPT3.5 + Expert Prompt                      & 16.5 (15) & 53.2 (15) & 40.5 (15) & 27.3 (15) & 31.3 (15) & 18.2 (15)\\
        GPT3.5 + \iclTargetAgnosticShort                 & 45.3 (15) & 96.0 (15) & 93.1 (15) & 90.4 (15) & 68.3 (15) & 63.6 (15) \\
        GPT3.5 + \iclTargetSpecificShort (Sent.)          & 58.1 (15) & 96.3 (15) & 83.2 (15) & 78.4 (15) & 73.6 (15) & 69.9 (15) \\
        GPT3.5 + \iclTargetSpecificShort (Doc.)           & \textbf{70.2} (5) & 95.4 (5) & 87.4 (1) & 84.6 (1) & 74.3 (15) & 60.3 (15) \\
        GPT3.5 + \iclTargetSpecificPlusWindowShort & 63.4 (15) & \textbf{96.6} (15) & \textbf{94.2} (15) & \textbf{92.4} (15) & \textbf{76.4} (15) & \textbf{72.6} (15) \\
        \midrule
        Zephyr-7B + \iclTargetSpecificPlusWindowShort                     & 42.1 (5)  & 80.0 (15) & 53.4 (1)  & 58.4 (10) & 43.9 (10) & 48.1 (15) \\
        \bottomrule
    \end{tabular}
    }
    \caption{TE normalization accuracy for English domains. The second number denotes the number of examples (sentences or documents) after which no performance increment was observed or the input length was exceeded. The best results using our proposed approach are in bold.}
    \label{tab:main_results}
\end{table*}

%% file: results_appendix.tex
\begin{table*}[!htpb]
    \centering
    \setlength\tabcolsep{6pt}
    \small
    \scalebox{1.0}{
    \begin{tabular}{lllllll}
        \toprule
        & \textbf{Ancient-Times} & \textbf{ECHR} & \textbf{ECJ} & \textbf{USC} & \textbf{i2b2} & \textbf{TempEval3} \\
        Domain & Wikipedia & Court & Court & Court & Clinical & News \\
        \midrule        
        GPT3.5 & 63.4 (15) & \textbf{96.6} (15) & \textbf{94.2} (15) & \textbf{92.4} (15) & \textbf{76.4} (15) & \textbf{72.6} (15) \\
        \midrule
        Zephyr-7B                     & 42.1 (5)  & 80.0 (15) & 53.4 (1)  & 58.4 (10) & 43.9 (10) & 48.1 (15) \\
        Llama2-7B & 8.0 (5) & 16.7 (15) & 5.1 (5) & 41.1 (15) & 2.6 (10) & 7.75 (10) \\
        Mistral-7B & 6.8 (5) & 27.3 (15) & 2.0 (1) & 35.3 (5) & 1.6 (5) & 7.0 (10) \\
        NeuralTrix & 4.7 (5) & 16.7 (15) & 6.7 (15) & 41.1 (15) & 2.5 (10) & 7.8 (15) \\
        Westlake-7B-v2 & 3.2 (5) & 16.7 (15) & 6.6 (15) & 41.1 (15) & 2.4 (10) & 7.8 (15) \\
        \bottomrule
    \end{tabular}
    }
    \caption{TE normalization accuracy with other language models. The second number denotes the number of examples (sentences or documents) after which no performance increment was observed or the input length was exceeded. All LLMs were prompted with our \iclTargetSpecificPlusWindowShort method}
    \label{tab:main_results}
\end{table*}